\documentclass[final]{cvpr}

\usepackage{times}
\usepackage{epsfig}
\usepackage{graphicx}
\usepackage{amsmath}
\usepackage{amssymb}
\usepackage{adjustbox}
\usepackage{soul}

\usepackage[dvipsnames]{xcolor}
\usepackage{subcaption}
\usepackage{booktabs}
\usepackage{multirow}
\usepackage{enumitem}
\usepackage{color, colortbl}
\usepackage{arydshln}
\usepackage{pifont}
\usepackage{appendix}

\newcommand{\cmpnas}{\textsc{Cmp-NAS}\xspace}
\newcommand{\hevs}{\textsc{HVS}\xspace}

\usepackage[pagebackref=true,breaklinks=true,colorlinks,bookmarks=false]{hyperref}

\def\cvprPaperID{2333} 
\def\confYear{CVPR 2021}

\DeclareMathOperator*{\argmax}{\arg\!\max}
\DeclareMathOperator*{\argmin}{\arg\!\min}

\setstcolor{red} 

\definecolor{Gray}{gray}{0.9}
\definecolor{amber(sae/ece)}{rgb}{1.0, 0.49, 0.0}
\definecolor{ao(english)}{rgb}{0.0, 0.5, 0.0}
\definecolor{cadmiumorange}{rgb}{0.93, 0.53, 0.18}
\definecolor{chocolate(web)}{rgb}{0.82, 0.41, 0.12}

\begin{document}

\title{Compatibility-aware Heterogeneous Visual Search}
\author{
Rahul Duggal\thanks{Currently at the Georgia Institute of Technology. Work conducted during an internship with Amazon AI.} \quad \ 
Hao Zhou  \quad \ 
Shuo Yang  \quad \ 
Yuanjun Xiong  \quad \\
Wei Xia\thanks{Corresponding author.}  \quad \
Zhuowen Tu  \quad \
Stefano Soatto  \\\\
AWS/Amazon AI \quad \\
{\tt\small  rduggal7@gatech.edu} \quad
{\tt\small \{zhouho, shuoy, yuanjx, wxia, ztu, soattos\}@amazon.com} 
}

\maketitle
\pagestyle{empty}
\thispagestyle{empty}
\begin{abstract}
{
We tackle the problem of visual search under resource constraints.
Existing systems use the same embedding model to compute representations (embeddings) for the query and gallery images.
Such systems inherently face a hard accuracy-efficiency trade-off: the embedding model needs to be large enough to ensure high accuracy, yet small enough to enable query-embedding computation on resource-constrained platforms. 
This trade-off could be mitigated if gallery embeddings are generated from a large model and query embeddings are extracted using a compact model.
The key to building such a system is to ensure representation compatibility between the query and gallery models. 
In this paper, we address two forms of compatibility: One enforced by modifying the parameters of each model that computes the embeddings. The other by modifying the architectures that compute the embeddings, leading to compatibility-aware neural architecture search  (\cmpnas).
We test \cmpnas on challenging retrieval tasks for fashion images (DeepFashion2), and face images (IJB-C). 
Compared to ordinary (homogeneous) visual search using the largest embedding model (paragon), \cmpnas achieves 80-fold and 23-fold cost reduction while maintaining accuracy within $0.3\%$ and $1.6\%$ of the paragon on DeepFashion2 and IJB-C respectively.
}
\end{abstract}

\section{Introduction}
\begin{figure}
    \centering
    \includegraphics[width=\linewidth]{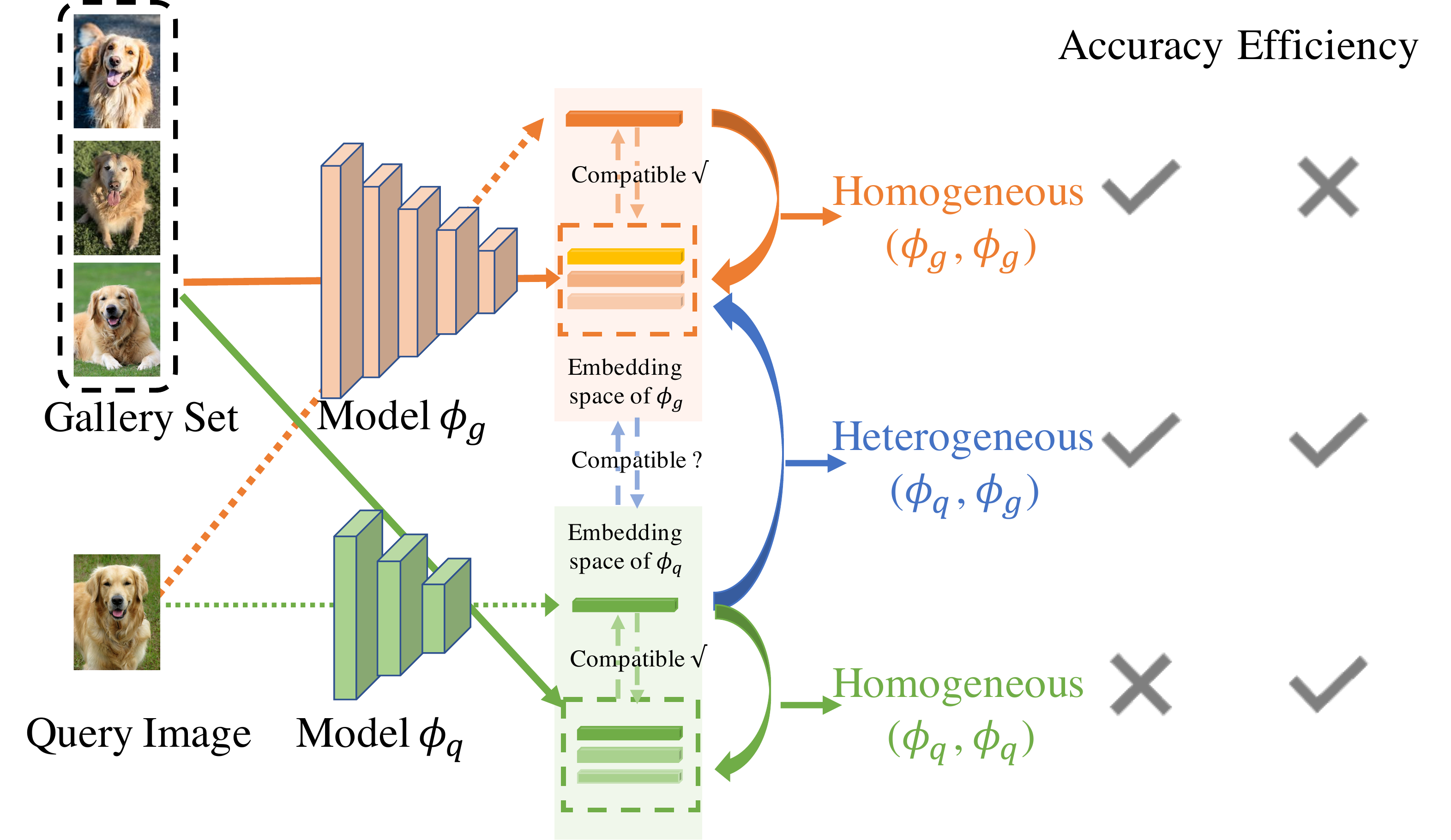}
    \caption{\small{
    Homogeneous visual search uses the same embedding model, either large (orange) to meet performance specifications, or small (green) to meet cost constraints, forcing a dichotomy. Heterogeneous Visual Search (blue) uses a large model to compute embeddings for the gallery, and a small model for the query images. This allows high efficiency without sacrificing accuracy,  provided that the green and orange embedding models are designed and trained to be {\em compatible}.}}
    \vspace{-0.5cm}
    \label{fig:crown_jewel}
\end{figure}

A visual search system in an ``open universe'' setting is often composed of a gallery model $\phi_g$ and a query model $\phi_q$, both mapping an input image to a vector representation known as \emph{embedding}.
The gallery model $\phi_g$ is typically used to map a set of gallery images onto their embedding vectors, a process known as indexing, while the query model extracts embeddings from query images to perform search against the indexed gallery. 
Most existing visual search approaches \cite{qayyum2017medical,razavian2016visual,xie2015image,babenko2015aggregating,tolias2015particular} use the same model architecture for both $\phi_q$ and $\phi_g$. We refer to this setup as \emph{homogeneous visual search}.  
An approach that uses different model architectures for $\phi_q$ and $\phi_g$ is referred to as \emph{heterogeneous visual search} (\hevs). 

The use of the same $\phi_g = \phi_q$ trivially ensures that gallery and query images are mapped to the same vector space where the search is conducted. However, this engenders a hard accuracy-efficiency trade-off (Fig.~\ref{fig:crown_jewel})---
choosing a large architecture $\phi_g$ for both query and gallery achieves high-accuracy at a loss of efficiency; choosing a small architecture $\phi_q$ improves efficiency to the detriment of accuracy, which is compounded since in practice, indexing only happens sporadically while querying is performed continuously. 
This leads to efficiency being driven mainly by the query model.
\hevs allows the use of a small model $\phi_q$ for querying, and a large model $\phi_g$ for indexing, partly mitigating the accuracy-complexity trade-off by enlarging the trade space.
The challenge in~\hevs is to ensure that $\phi_g$ and $\phi_q$ live in the same metric (vector) space. 
This can be done for given architectures $\phi_g, \phi_q$, by training the weights so the resulting embeddings are metrically compatible \cite{shen2020towards}. However, one can also enlarge the trade space by including the architecture in the design of metrically compatible models. Typically, $\phi_g$ is chosen to match the best current state-of-the-art (paragon) while the designer can search among query architectures $\phi_q$ to maximize efficiency while ensuring that performance remains close to the paragon.

\begin{figure}
    \centering
    \includegraphics[width=\linewidth]{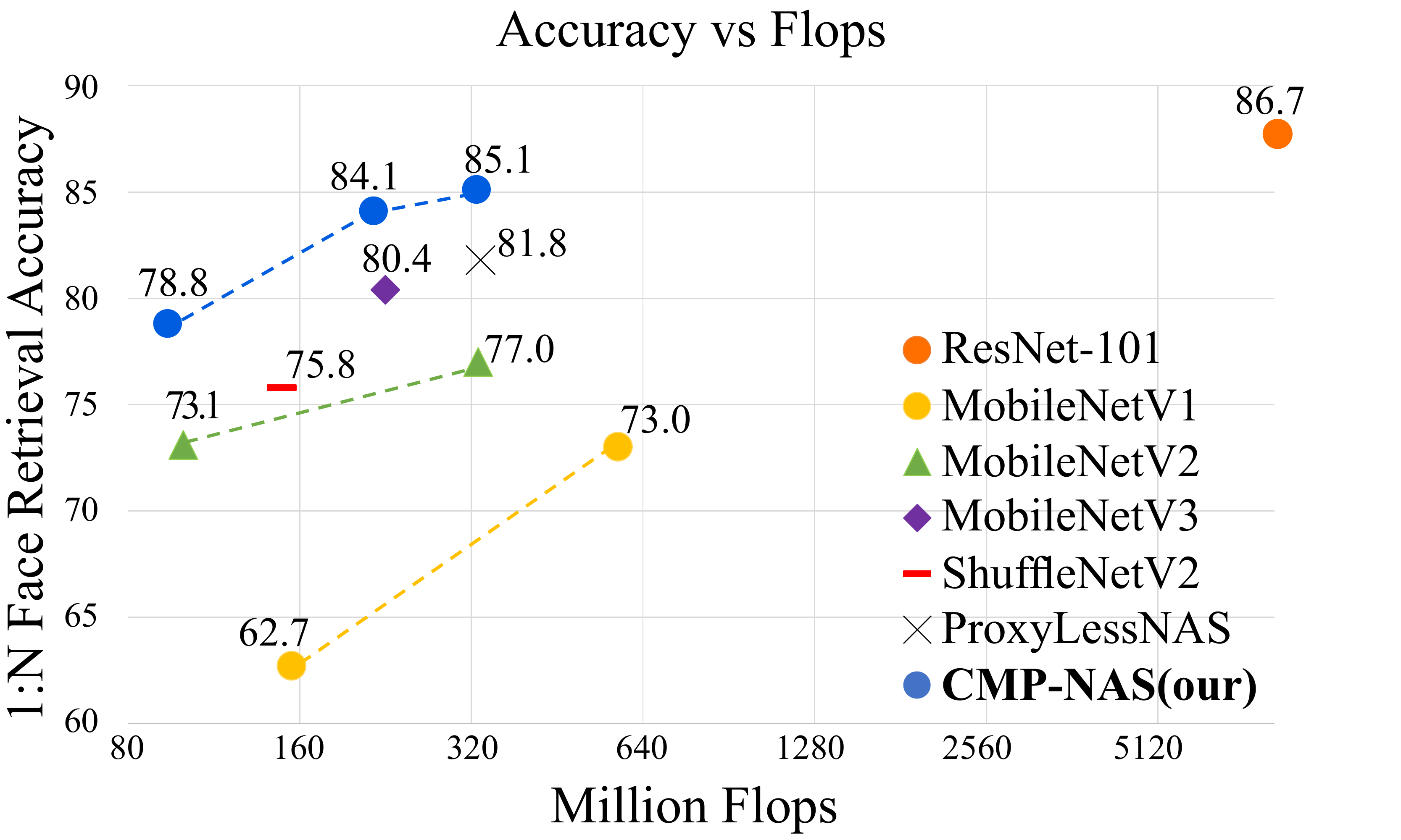}
    \caption{\small{The trade-off between accuracy and efficiency for a heterogeneous system performing 1:N Face retrieval on DeepFashion2. We use a ResNet-101 as the gallery model and compare different architectures as query models.  For MobileNetV1 and V2, we provide results with width $0.5\times$ and $1\times$}.}
    \vspace{-0.5cm}
    \label{fig:crown_jewel_performance}
\end{figure}

In this work, we pursue compatibility by optimizing both the model parameters (weights) as well as the model architecture. 
We show that (1) weight inheritance~\cite{rethinking_pruning_value} and (2) backward-compatible training (BCT)~\cite{shen2020towards} can achieve compatibility through weight optimization. 
Among these, the latter is more general in that it works with {\em arbitrary} embedding functions $\phi_g$ and $\phi_q$. 
We expand beyond BCT to neural architecture search (NAS)~\cite{NAS_RL_Quoc,proxylessnas,NAS_SinglePath,MnasNet} with our proposed compatibility-aware NAS (\cmpnas) strategy that searches for a query model $\phi_q$ that is maximally efficient while being compatible with $\phi_g$. We hypothesize that~\cmpnas can simultaneously find the architecture of query model and its weights that achieve efficiency similar to that of the smallest (query) model, and accuracy close to that of the paragon (gallery model). Indeed the results in Fig.~\ref{fig:crown_jewel_performance} shows that~\cmpnas outperforms all of the state-of-the-art off-the-shelf architectures designed for mobile platforms with resource constraints. Compared with paragon (state-of-the-art high-compute homogeneous visual search) methods, \hevs reduce query model flops by $23\times$ with only $1.6\%$ in loss of search accuracy for the task of face retrieval.

Our {\em contributions} can be summarized as follows: 1) we demonstrate that an \hevs system allows to better trade off accuracy and complexity, by optimizing over both model parameters and architecture. 2) We propose a novel~\cmpnas method combining weight-based compatibility with a novel reward function to achieve compatibility-aware architecture search for~\hevs. 3) We show that our \cmpnas can reduce model complexity many-fold with only a marginal drop in accuracy. For instance, we achieve $23\times$ reduction in flops with only $1.6\%$ drop in retrieval accuracy on face retrieval using standard benchmarks.

\section{Related Work}
\label{sec:related_works}

\textbf{Visual search:} 
Most prior visual search systems construct embedding vectors either by aggregating hand-crafted features~\cite{wengert2011bag,park2002fast,siagian2007rapid,arandjelovic2012three,zheng2014packing,zhou2012scalar}, or through feature maps extracted from a convolutional neural network~\cite{razavian2016visual,zheng2015query,xie2015image,uijlings2013selective,razavian2016visual,babenko2015aggregating,kalantidis2016cross,tolias2015particular,radenovic2018fine}. 
The latter, being more prevalent in recent times, differ from us in that they follow the homogeneous visual search setting and suffer from a hard accuracy and efficiency trade-off. 
Recently, \cite{budnik2020asymmetric} discusses the asymmetric testing task which is similar to our heterogeneous setting. However, their method is unable to ensure that the heterogeneous accuracy supersedes the homogeneous one (compatibility rule in Sec.~\ref{subsec:NotationCompatibility}). Such a system is not practically useful since the homogeneous deployment achieves both a higher accuracy and a higher efficiency.

\textbf{Cross-model compatibility:} 
The broad goal of this area is to ensure embeddings generated by different models are compatible. 
Some recent works ensure cross-model compatibility by learning transformation functions from the query embedding space to the gallery one \cite{wang2020unified,chen2019r3,hu2019towards}. 
Different from these works, our approach directly optimizes the query model such that its metric space aligns with that of the gallery.
This leads to more flexibility in designing the query model and allows us to introduce architecture search in the metric space alignment process.
Our idea of model compatibility, as metric space alignment, is similar to the one in backward-compatible training (BCT) \cite{shen2020towards}. However,
\cite{shen2020towards} only considers compatibility through model weights, whereas, we generalize this concept to the model architecture. Additionally, \cite{shen2020towards} targets for compatibility between an updated model and its previous (less powerful) version, the application scenarios of which are different from this work.

\textbf{Architecture Optimization:}
Recent progress demonstrates the advantages of automated architecture design over manual design through techniques such as neural architecture search (NAS) \cite{NAS_RL_Quoc,MnasNet,NAS_SinglePath,proxylessnas}.
Most existing NAS algorithms however, search for architectures that achieve the best accuracy when used independently. In contrast, our task necessitates a deployment scenario with two models: one for processing the query images and another for processing the gallery.
Recently, \cite{SearchDistill, BlockWiselySearch} propose to use a large teacher model to guide the architecture search process for a smaller student which is essentially knowledge distillation in architecture space. However, our experiments show that knowledge distillation cannot guarantee compatibility and thus these methods may not succeed in optimizing the architecture in that aspect. To the best of our knowledge, \cmpnas is the first to consider the notion of compatibility during architecture optimization.
\section{Methodology}

We use $\phi$ to denote an embedding model in a visual search system and $\kappa$ to denote the classifier that is used to train $\phi$. We further assume $\phi$ is determined by its architecture $a$ and weights $w$.
For our visual search system, a gallery model is first trained on a training set $\mathcal{T}$ and then used to map each image $x$ in the gallery set $\mathcal{G}$ onto an embedding vector $\phi_g(x) \in \mathbb{R}^{K}$. Note this mapping process only uses the embedding portion $\phi_g$.
During test time, we use the query model (trained previously on $\mathcal{T}$) to map the query image $x'$ onto an embedding vector $\phi_q(x') \in \mathbb{R}^{K}$.
The closest match is then obtained through a nearest neighbor search in the embedding space.
Typically, visual search accuracy is measured through some metric, such as top-10 accuracy, which we denote by $M(\phi_q, \phi_g; \mathcal{Q}, \mathcal{G})$.  This is calculated by processing query image set $\mathcal{Q}$ with $\phi_q$ and processing the gallery set $\mathcal{G}$ with $\phi_g$. For simplification, we omit the image sets and adopt the notation $M(\phi_q, \phi_g)$ to denote our accuracy metric.

\subsection{Homogeneous \vs heterogeneous visual search}
Assuming $\phi_q$ and $\phi_g$ are different models and $\phi_q$ is smaller than $\phi_g$, we define two kinds of visual search:
\begin{itemize}
    \item \textbf{Homogeneous visual search} uses the same embedding model to process the gallery and query images, and is denoted by ($\phi_q$, $\phi_q$) or ($\phi_g$, $\phi_g$).
    \item \textbf{Heterogeneous visual search} uses different embedding models to process the query and gallery images, respectively, and is denoted by ($\phi_q$, $\phi_g$).
\end{itemize}

\begin{figure}
    \centering
    \begin{subfigure}{0.495\linewidth}
    \centering
    \includegraphics[width=\linewidth]{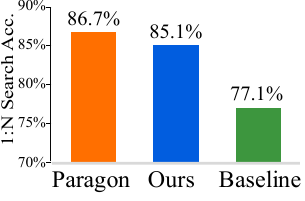}
    \caption{\small{Accuracy on IJB-C.}}
    \end{subfigure}
    \begin{subfigure}{0.495\linewidth}
    \centering
    \includegraphics[width=\linewidth]{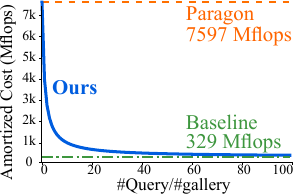}
    \caption{\small{Amortized cost analysis}}
    \end{subfigure}
    \caption{\small{
    Accuracy-efficiency trade-off for visual search. In (a) we compare the 1:N face retrieval accuracy (TPIR@FPIR=$10^{-1}$) on IJB-C. We denote the homogeneous system with ResNet-101 and MobileNetV2 as the paragon and baseline respectively. In (b) we observe that, as the size of the query set increases, the complexity of our heterogeneous system converges to that of the baseline.}}
    \label{fig:face_complex}
    \vspace{-5mm}
\end{figure}

We illustrate the accuracy-efficiency trade-off faced by visual search systems in Fig.~\ref{fig:face_complex}. A homogeneous system with a larger embedding model (\eg ResNet-101 \cite{he2016deep}, denoted as paragon) achieves a higher accuracy due to better embeddings (orange bar in Fig.~\ref{fig:face_complex}(a)) but also consumes more flops during query time (orange line in Fig.~\ref{fig:face_complex}(b)). On the other hand, a smaller embedding model (\eg MobileNetV2 \cite{MobileNetV2}, denoted as baseline) in the homogeneous setting achieves the opposite end of the trade-off (green bar and line in Fig.~\ref{fig:face_complex}(a),(b)). Our heterogeneous system (blue bar and line in Fig.~\ref{fig:face_complex}(a),(b)) achieves accuracy within $1.6\%$ of the paragon and efficiency of the baseline.

When computing the cost of a visual search method, one has to take into account both the cost of indexing, which happens sporadically, and the cost of querying, which occurs continuously. While large, the indexing cost is amortized through the lifetime of the system. To capture both, in Fig.~\ref{fig:face_complex}(b) we report the amortized cost of embedding the query and gallery images, as a function of the ratio of queries to gallery images processed. In most practical systems, the number of queries exceeds the number of indexed images by orders of magnitude, so the relevant cost is the asymptote, but we report the entire curve for completeness. The initial condition for that curve is the cost of the paragon. Our goal is to design a system that has a cost approaching the asymptote (b), with performance approaching the paragon (a).

\subsubsection{Notion of Compatibility}
\label{subsec:NotationCompatibility}

A key requirement of a heterogeneous system is that the query and gallery models should be compatible. We define this notion through the \textbf{compatibility rule} which states that:
\begin{itemize}
    \item[] A {\em smaller} model $\phi_q$ is compatible with a {\em larger} model $\phi_g$ if it satisfies the inequality $M(\phi_q,\phi_g) > M(\phi_q,\phi_q)$.
\end{itemize}
We note that satisfying this rule is a necessary condition for heterogeneous search. 
A heterogeneous system violating this condition, 
\ie $M(\phi_q,\phi_g) < M(\phi_q,\phi_q)$, is not practically useful since the homogeneous system $M(\phi_q,\phi_q)$ achieves both higher efficiency and accuracy.
Additionally, a practically useful heterogeneous system should also satisfy $M(\phi_q,\phi_g) \approx M(\phi_g,\phi_g)$.
In subsequent sections, we study how to achieve both these goals through weight and architecture compatibility.

\begin{figure*}
    \centering
    \begin{subfigure}{0.3\linewidth}
    \centering
    \includegraphics[width=0.95\linewidth]{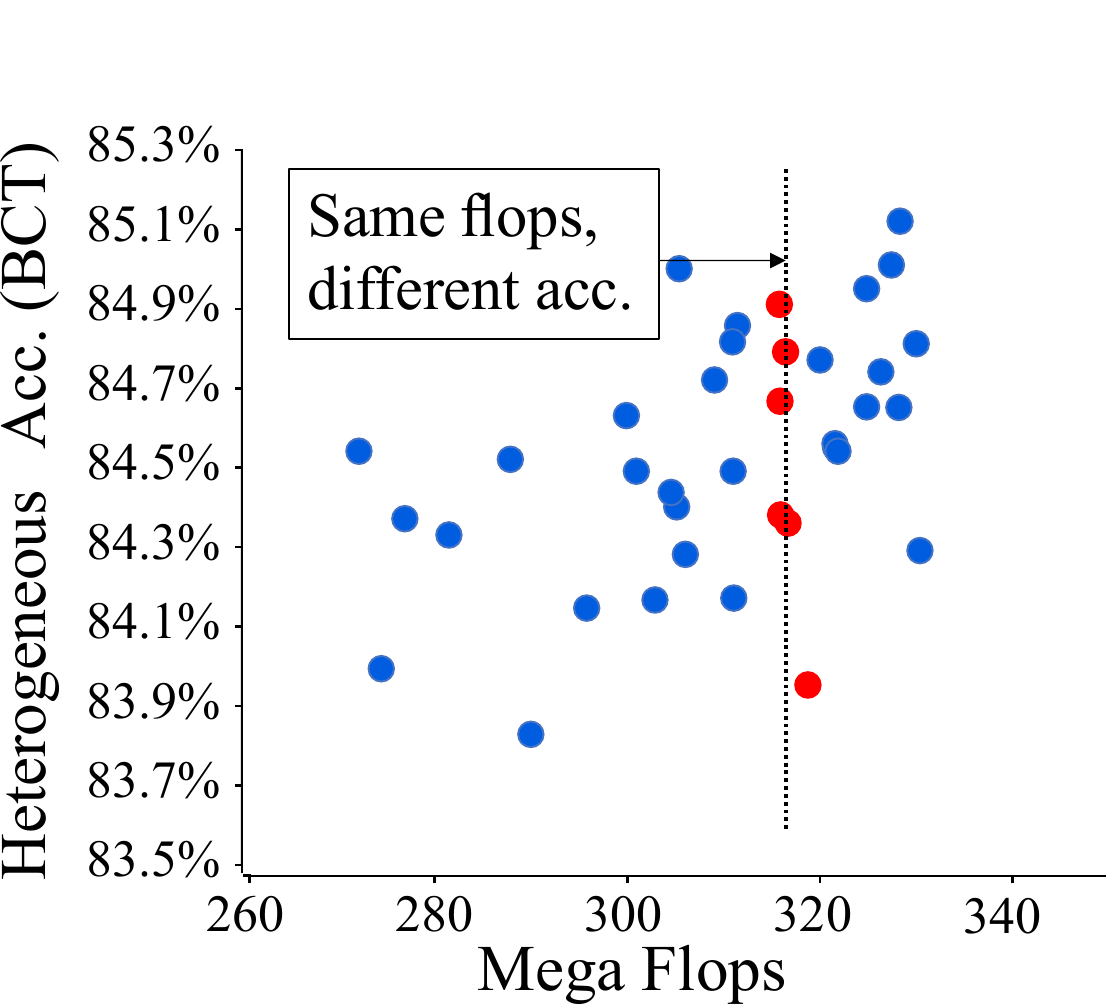}
    \caption{}
    \end{subfigure}
    \begin{subfigure}{0.3\linewidth}
    \centering
    \includegraphics[width=0.95\linewidth]{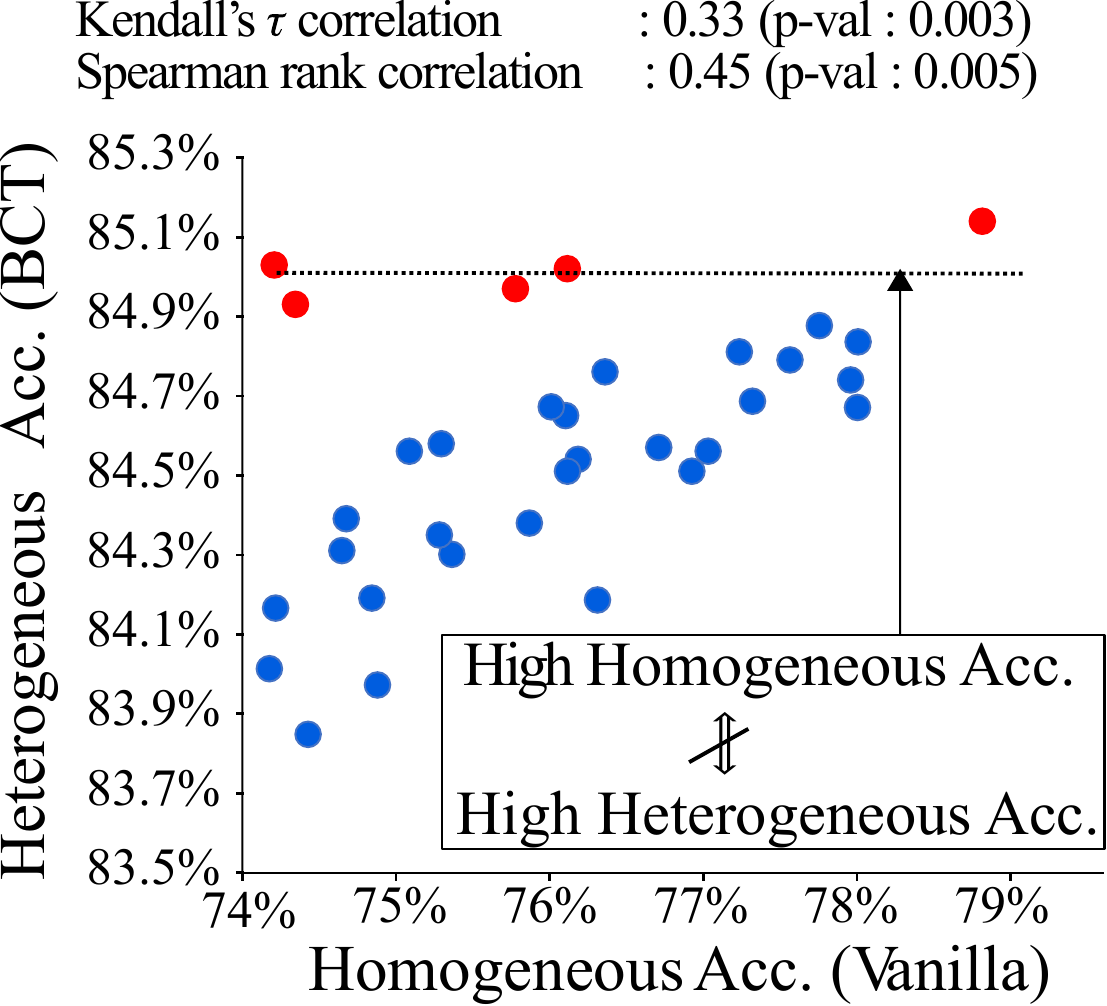}
    \caption{}
    \end{subfigure}
     \begin{subfigure}{0.3\linewidth}
    \centering
    \includegraphics[width=0.95\linewidth]{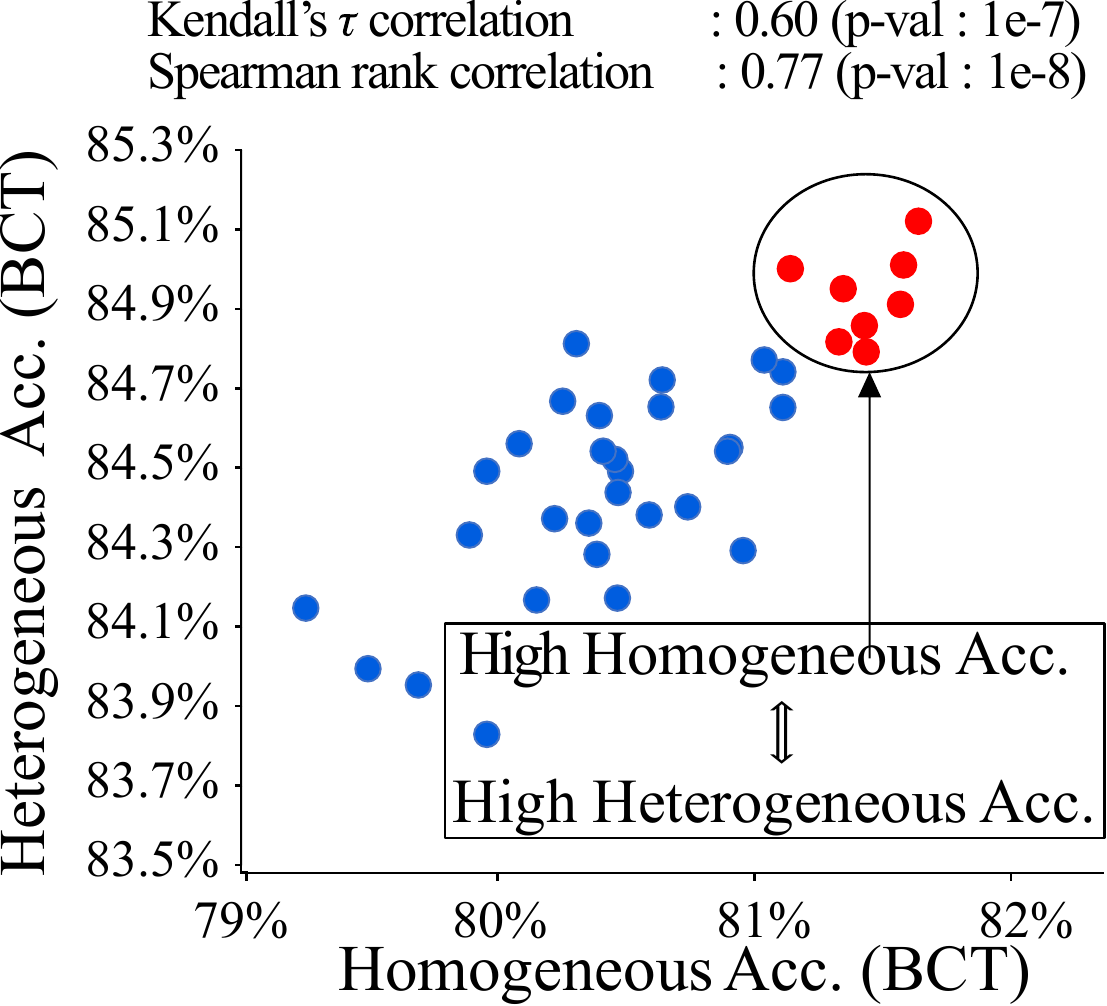}
    \caption{}
    \end{subfigure}
    \caption{\small{NAS Motivation. We randomly sample 40 architectures from the ShuffleNet search space of \cite{NAS_SinglePath} and train them from scratch. Observe that (a) Architectures with same flops (shown with red circles) can have different heterogeneous accuracies proving that architecture has a measurable impact on compatibility. (b) Architectures (shown in red) achieving the highest heterogeneous accuracy with BCT training are not the ones achieving the highest homogeneous accuracy with vanilla training. This means that traditional NAS (which optimizes for homogeneous accuracy while using vanilla training) may fail to find the most compatible models. (c) When trained with BCT, the architectures achieving the highest heterogeneous accuracy also achieve the highest homogeneous accuracy. This means simply equipping traditional NAS with BCT will aid the search for compatible architectures.}}
    \label{fig:nas_motivation}
    \vspace{-5mm}
\end{figure*}

\subsection{Compatibility for Heterogeneous Models}
\label{subsec:representational_compatibility}
In this section, we discuss different ways to obtain compatible query and gallery models $\phi_q$, $\phi_g$ that satisfy the compatibility rule. While a general treatment may optimize $\phi_q$ and $\phi_g$ jointly, in this paper, we consider the simpler case when $\phi_g$ is fixed to a standard large model (ResNet-101) while we optimize the query model $\phi_q$. 
For the subsequent discussion, we assume the gallery model $\phi_g$ has an architecture $a_g$ and is parameterized by weights $w_g$.
Corresponding quantities for the query model are $\phi_q$ with architecture $a_q$ and parameterized by weights $w_{q}$. 
To train the query and gallery models $\phi_q$, $\phi_g$ we use the classification-based training \cite{shen2020towards,Zhai2019ClassificationIA} with the query and gallery classifiers denoted by $\kappa_q$ and $\kappa_g$ respectively.
In what follows, we discuss two levels of compatibility---weight level and architecture level.

\subsubsection{Weight-level compatibility}


Given the gallery model $\phi_g$ and its classifier $\kappa_g$, weight-level compatibility aims to learn the weights $w_q$ of query model $\phi_q$ such that the compatibility rule is satisfied. To this end, the optimal query weights $w_{q}^*$, and its corresponding classifier $\kappa_q^*$ can be learned by minimizing a composite loss over the training set $\mathcal{T}$.
\begin{multline}
      w_{q}^*, \kappa_q^* = \argmin_{w_{q}, \kappa_q} \{\lambda_1 \mathcal{L}_1( w_{q}, \kappa_q; \mathcal{T}) + \\ \lambda_2  \mathcal{L}_2( w_{q}, \kappa_q, w_{g}, \kappa_g; \mathcal{T})\},
       \label{eq:compat_training}
\end{multline}
where $\mathcal{L}_1$ is a classification loss such as the Cosine margin \cite{wang2018cosface}, Norm-Softmax \cite{norm_softmax} and $ \mathcal{L}_2$ is the additional term which promotes compatibility.
We consider four training methods which can be described using Eq.\ref{eq:compat_training} as follows:
\begin{enumerate}[itemsep=0.1mm]
    \item Vanilla training: Considers $\lambda_2=0$.
    \item Knowledge Distillation \cite{hinton2015distilling}: $\mathcal{L}_2$ is the temperature smoothed cross-entropy loss between the logits of query and gallery model.
    \item Fine-tuning: Initializes $w_{q}$ using $w_{g}$ and $\kappa_q$ using $\kappa_g$ and considers $\lambda_2=0$.
    \item Backward-compatible training (BCT) \cite{shen2020towards}: Uses $\mathcal{L}_2 = \mathcal{L}_1(w_{q}, \kappa_g; \mathcal{T})$. This ensures that the query embedding model learns a representation that is compatible with the gallery classifier.
\end{enumerate}

We compare these methods in Tab.~\ref{tab:compatibility_comparison}, and find that only the last two succeed in ensuring compatibility. Among these two, fine-tuning is more restrictive since it makes a stronger assumption about the query architecture---it requires the query model to have a similar network structure, kernel size, layer configuration as the gallery model. In contrast, BCT poses no such restriction and can be used to train any query architecture. Thus we use \cite{shen2020towards} as our default method to ensure weight-level compatibility. Recently \cite{budnik2020asymmetric} proposed to learn the weights of a query model by minimizing the $L_2$ distance between query and gallery embeddings, however, both \cite{budnik2020asymmetric} and \cite{shen2020towards} observe that the resulting query model does not satisfy the compatibility rule.

\subsubsection{Architecture-level compatibility}
\label{subsec:Architecture optimization}
Given the gallery model $\phi_g$ and classifier $\kappa_g$, the problem of architecture-level compatibility aims to search for an architecture $a_q$ for the query model $\phi_q$ that is most compatible with a fixed gallery model. The need of architecture level compatibility is motivated by two questions:
\begin{enumerate}[itemsep=0.1mm]
    \item[Q1] How much does architecture impact compatibility? 
    \item[Q2] Can traditional NAS find compatible architectures?
\end{enumerate}
To answer these questions, we randomly sample 40 architectures from the ShuffleNet search space \cite{NAS_SinglePath}, with each having roughly $300$ Million flops. 
\begin{enumerate}[itemsep=0.1mm]
    \item[A1] We train these architectures with BCT ($\lambda_1=1, \lambda_2=1$ in Eq.~\ref{eq:compat_training}) and plot the heterogeneous accuracy \vs flops in Fig.~\ref{fig:nas_motivation}(a). There are two observations: (1) heterogeneous accuracy is not correlated with flops and (2) architectures with similar flops can achieve different accuracy, which indicates architecture indeed has a measurable impact on accuracy.
    \item[A2] We plot the homogeneous accuracy of models with vanilla training (target of traditional NAS) \vs heterogeneous accuracy of the same models trained with BCT (our target) in Fig.~\ref{fig:nas_motivation}(b). We observe that: (1) The correlation between the two accuracy is low and (2) The architectures with the highest heterogeneous accuracy are not those with highest homogeneous accuracy. This indicates traditional NAS may not be  successful in searching for compatible architectures. 
\end{enumerate}
We further investigate the correlation between homogeneous (with BCT) and heterogeneous accuracy (with BCT) in Fig.~\ref{fig:nas_motivation}(c) and discover that the correlation of these two accuracies is much higher than that in Fig.~\ref{fig:nas_motivation}(b). This offers a key insight that equipping traditional NAS with BCT may help in searching compatible architectures.

\noindent\textbf{Architecture optimization with \cmpnas }
Based on the intuition developed previously, we develop \cmpnas using the following notation.
Denote by $\phi_q(a_q,w_q)$ a candidate query embedding model with architecture $a_q$ and weights $w_q$. We further denote $\kappa_q$ as its corresponding classifier.
With \cmpnas, we solve a two-step optimization problem where the first step amounts to learning the best set of weights--- $w_{q}^*$ (for the embedding model $\phi_q$) and $\kappa_q^*$  (for the common classifier)--by minimizing a classification loss $\mathcal{L}$ over the training set $\mathcal{T}$ as below

\begin{multline}
    w_{q}^*,  \kappa_q^* = \argmin_{w_{q}, \kappa_q} \{ \lambda_1 \mathcal{L}(\phi_q(a_q,w_{q}),\kappa_q; \mathcal{T}) \\ +  \lambda_2 \mathcal{L}(\phi_q(a_q,w_{q}),\kappa_g; \mathcal{T}) \},
    \label{eq:NAS_supernet_train_with_BCT}
\end{multline}
where $\mathcal{L}$ can be any classification loss such as Cosine margin \cite{wang2018cosface}, Norm-Softmax \cite{norm_softmax}.
Similar to BCT, the second term $\mathcal{L}(\phi_q(a_q,w_q),\kappa_g; \mathcal{T})$ ensures that the candidate query embedding model $\phi_q(a_q,w_q^*)$ learns a representation that is compatible with the gallery classifier.

Using weights $w_q^*$ and $\kappa_q^*$ from above, the second step amounts to finding the best query architecture $a_q^*$ in a search space $\Omega$, by maximizing a reward $\mathcal{R}$ evaluated on the validation set as below
\begin{equation}
    a_q^* = \argmax_{a_q \in \Omega} \mathcal{R} \left(\phi_q(a_q,w_{q}^*), \kappa_q^* \right).
    \label{eq:NAS_search_without_BCT}
\end{equation}
We consider three candidate rewards presented in Tab~\ref{tab:reward}. Similar to traditional NAS, homogeneous accuracy $M(\phi_q(a_q, w_q), \phi_q(a_q, w_q))$ is our baseline reward $\mathcal{R}_1$. Recall that however, we are interested in searching for the architecture which achieves the best heterogeneous accuracy. With this aim, we design rewards $\mathcal{R}_2$ and $\mathcal{R}_3$ which include the heterogeneous accuracy in their formulation.

\begin{table}[]
\centering
\small
\begin{tabular}{@{}ll@{}}
\toprule
Reward & Formulation  \\ \midrule
$\mathcal{R}_1$ & $M(\phi_q(a_q, w_q), \phi_q(a_q, w_q)$\\
$\mathcal{R}_2$ & $M(\phi_q(a_q, w_q),\phi_g)$ \\
$\mathcal{R}_3$ & $M(\phi_q(a_q, w_q), \phi_q(a_q, w_q)) \times M(\phi_q(a_q, w_q), \phi_g)$\\ \bottomrule
\end{tabular}
\caption{\small{Different rewards considered with \cmpnas. The rewards $\mathcal{R}_1$, $\mathcal{R}_2$ prioritize either the symmetric or asymmetric accuracy while ignoring the other. $\mathcal{R}_3$ prioritizes \textit{both} accuracies and consistently outperforms other rewards.}}
\label{tab:reward}
\vspace{-3mm}
\end{table}

Our \cmpnas formulation in Eq.~\ref{eq:NAS_supernet_train_with_BCT} and rewards in Tab.~\ref{tab:reward} is general and can work with any NAS method. For demonstration, we test our idea with the single path one shot NAS \cite{NAS_SinglePath} and consists of the following two components:

\noindent \textbf{Search Space:} Similar to popular weight sharing methods \cite{proxylessnas,NAS_SinglePath,DARTS}, the search space of our query model consists of a shufflenet-based super-network. The super-network consists of $20$ sequentially stacked choice blocks. Each choice block can select one of four operations: $k\times k$ convolutional blocks ($k\in {3,5,7}$) inspired by ShuffleNetV2 \cite{ma2018shufflenet} and a $3\times3$ Xception \cite{chollet2017xception} inspired convolutional block. Additionally, each choice block can also select from 10 different channel choices $0.2-2.0 \times$.  
During training we use the loss formulation in Eq.~\ref{eq:NAS_supernet_train_with_BCT} to train this super-network whereby, in each batch a new architecture is sampled uniformly \cite{NAS_SinglePath} and only the weights corresponding to it are updated. 

\noindent \textbf{Search Strategy:} To search for the most compatible architecture, \cmpnas uses evolutionary search \cite{NAS_SinglePath} fitted with the different rewards outlined in Tab.~\ref{tab:reward}. The search is fast because each architecture inherits the weights from the super-network. In the end, we obtain the five best architectures and re-train them from scratch with BCT.

\section{Experiments}
We evaluate the efficacy of our heterogeneous system on two tasks: face retrieval, as it is one of the ``open-universe'' problems with the largest publicly available datasets; and fashion retrieval which necessitates an open-set treatment due to the constant evolution of fashion items. We use face retrieval as the main benchmark for our ablation studies.

\subsection{Datasets, Metrics and Gallery Model}
\textbf{Face Retrieval:} We use the IMDB-Face dataset \cite{IMDB} to train the embedding model for the face retrieval task. The IMDB-Face dataset contains over 1.7M images of about 59k celebrities. If not otherwise specified, we use $95\%$ of the data as training set to train our embedding model and use the remaining $5\%$ as a validation dataset to compute the rewards for architecture search. For testing, we use the widely used IJB-C face recognition benchmark dataset \cite{IJBC}. The performance is evaluated using the true positive identification rate at a false positive identification rate of $10^{-1}$ (TPIR@FPIR=$10^{-1}$). Throughout the evaluation, we use a ResNet-101 as the fixed gallery model $\phi_g$.

\textbf{Fashion Retrieval:}
We evaluate the proposed method on Commercial-Consumer Clothes Retrieval task on DeepFashion2 dataset \cite{Deepfashion2}. It contains 337K commercial-consumer clothes pairs in the training set, from which $90\%$ of the data is used for training the embedding and the rest $10\%$ is used for computing the rewards in architecture search. We report the test accuracy using Top-10 retrieval accuracy on the original validation set, which contains 10,844 consumer images with 12,377 query items, and 21,309 commercial images with 36,961 items in the gallery. ResNet-101 is used as the fixed gallery model $\phi_g$.

\subsection{Implementation Details} 
Our query and gallery models take a $112\times112$ image as input and output an embedding vector of $128$ dimensions.

\noindent \textbf{Face retrieval:} We use mis-alignment and color distortion for data augmentation \cite{shen2020towards}. Following recent state-of-the-art \cite{wang2018cosface}, we train our gallery ResNet-101 model using the cosine margin loss \cite{wang2018cosface} with margin set to $0.4$. We use the SGD optimizer with weight decay $5\times10^{-4}$. The initial learning rate is set to $0.1$ which decreases to $0.01$, $0.001$ and $0.0001$ after $8$, $12$ and $14$ epochs. Our gallery model is trained for $16$ epochs with a batch size $320$. We train the query models for 32 epochs with a cosine learning rate decay schedule \cite{Cosine_Annealing}. The initial learning rate is set to $1.3$ all query models except MobileNetV1($1\times$) which uses $0.1$.

\noindent \textbf{Fashion retrieval:} The original fashion retrieval task with DeepFashion2 \cite{Deepfashion2} requires to first detect and then retrieve fashion items. Since we only tackle the retrieval task, we construct our retrieval-only dataset by using ground truth bounding box annotations to extract the fashion items. To train the gallery model, we follow \cite{Zhai2019ClassificationIA} in using normalized cross entropy loss with temperature $0.5$. The gallery model is trained for 40 epochs using an initial learning rate of $3.0$ with cosine decay. The weight decay is set to $10^{-4}$. Our query models are trained with BCT for 80 epochs using an initial learning rate of $10$ with cosine decay.

\noindent \textbf{Runtime:}  On a system containing 8 Tesla V100 GPUs, the entire pipeline for the face (and fashion) retrieval takes roughly 100 (45) hours. This includes roughly 8 (8) hours to train the gallery model, 32 (14) hours to train the query super-network, 48 (20) hours for evolutionary search and 2 (2) hours to train the final query architecture.

For additional implementation details specific to \cmpnas, please refer to the supplementary material.

\begin{table}[]
    \centering
    \begin{adjustbox}{width=\linewidth,center}
    \begin{tabular}{l ll rr}
    \toprule
    & Gallery  & Query & Acc.        &  Query    \\
    & Model    & Model & (\%)   & Flops (M) \\
    \midrule
    Paragon &ResNet-101 & ResNet-101 & 86.7  & 7597 \\
    \cdashline{1-5}[.7pt/1.5pt]\noalign{\vskip 0.15em}
    \multirow{1}{*}{Proposed}   & ResNet-101 & \cmpnas & 85.1  & 327 \\
     \cdashline{1-5}[.7pt/1.5pt]\noalign{\vskip 0.15em}
    Baseline &MobileNetV2 & MobileNetV2 & 77.1  & 329\\
    \bottomrule 
    \end{tabular}

    \end{adjustbox}
    \caption{\small{Comparison with baseline and paragon for 1:N face retrieval on IJB-C. Accuracy is reported as TPIR(\%)@FPIR=$10^{-1}$. All the models except the paragon are trained with BCT loss using ResNet-101 as the ``teacher''.}}
    \label{tab:Face_BP}
    \vspace{-3mm}
\end{table}

\begin{table}[]
    \centering
    \begin{adjustbox}{width=\linewidth,center}
    \begin{tabular}{l ll rr}
    \toprule
    & Gallery  & Query & Acc.    &  Query    \\
    & Model    & Model & (\%)   & Flops (M) \\ \midrule
    Paragon &ResNet-101 & ResNet-101 & 65.2 & 7597 \\
    \cdashline{1-5}[.7pt/1.5pt]\noalign{\vskip 0.15em}
    \multirow{1}{*}{Proposed}   & ResNet-101 & \cmpnas & 64.9 &211 \\
    \cdashline{1-5}[.7pt/1.5pt]\noalign{\vskip 0.15em}
    Baseline &MobileNetV3 & MobileNetV3 & 62.7 & 226\\
    \bottomrule 
    \end{tabular}
    \end{adjustbox}
    \caption{\small{Comparison with baseline and paragon for fashion retrieval on DeepFashion2. Accuracy is reported as Top-10 retrieval accuracy. All the models except the paragon are trained with BCT loss using ResNet-101 as the ``teacher''.}}
    \label{tab:Fashion_BP}
    \vspace{-3mm}
\end{table}

\subsection{Baseline and paragon for visual search}

Since gallery features can be pre-computed and there is no computational constraint on the gallery side, we fix the gallery model to a ResNet-101.
In terms of visual search accuracy, the paragon is achieved by the (ResNet-101, ResNet-101) system on both the face and fashion retrieval tasks. 
On the other hand, we use (MobileNetV2, MobileNetV2) and (MobileNetV3, MobileNetV3) as the baseline for face and fashion retrieval respectively, since they achieve the highest accuracy among the MobileNet family.

Tab.~\ref{tab:Face_BP} shows almost a $10\%$ gap in accuracy between the paragon and baseline for face retrieval. 
On the other end, the baseline consumes $23\times$  fewer query flops than the paragon.
This establishes the goal of our heterogeneous system: To achieve accuracy similar to the paragon while consuming query flops similar to the baseline. 
Indeed, the middle rows of Tab.~\ref{tab:Face_BP} shows this goal is achieved by our proposed heterogeneous system (ResNet-101, \cmpnas) which consumes similar query flops as the baseline with only $1.6\%$ accuracy drop compared to the paragon.
Tab.~\ref{tab:Fashion_BP} reveals a similar observation on DeepFashion2.

\subsection{Dissecting the performance of \cmpnas}
In this subsection, we break down the accuracy achieved by our heterogeneous system in terms of the improvement due to (1) weights and (2) architecture compatibility.

\begin{figure}
    \centering
    \begin{subfigure}{\linewidth}
    \centering
    \includegraphics[width=0.9\linewidth]{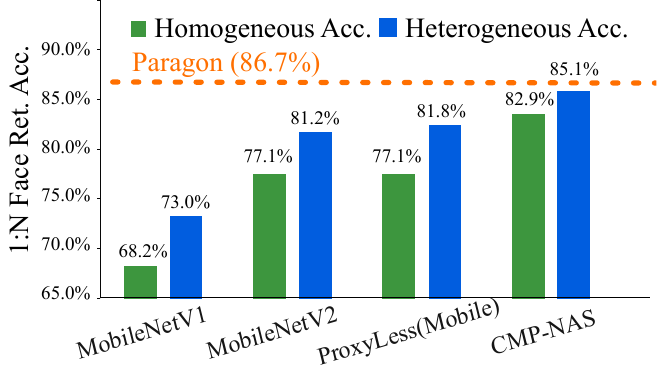}
    \caption{\small{1:N Face retrieval accuracy (TPIR@FPIR=$10^{-1}$) on IJB-C. }}
    \vspace{5mm}
    \end{subfigure}
    \begin{subfigure}{\linewidth}
    \centering
    \includegraphics[width=0.9\linewidth]{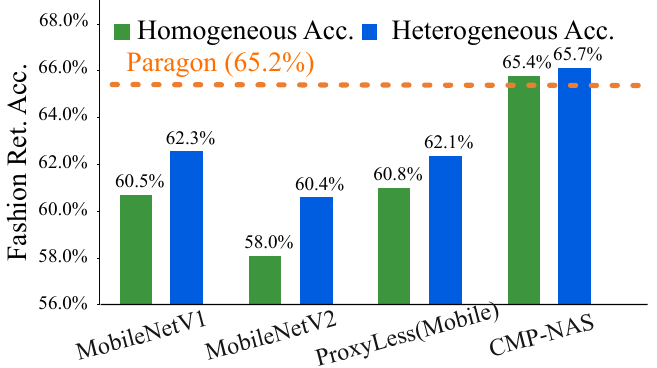}
    \caption{\small{
    Fashion retrieval accuracy (top-10) on DeepFashion2.}}
    \end{subfigure}
    \caption{\small{Evaluating the heterogeneous and homogeneous search accuracy for face and fashion retrieval tasks using different query models. \cmpnas outperforms other baselines and achieves accuracy close to the paragon. 
    }}
    \label{fig:acc_break_down}
    \vspace{-5mm}
\end{figure}

\noindent \textbf{Improvement due to weight-compatibility:}
To observe the improvement due to weight compatibility, Fig.~\ref{fig:acc_break_down} (a),(b) shows the homogeneous and heterogeneous accuracy obtained by three state-of-the-art query models with \textit{static} architectures in the 300 Million flops range trained with BCT (Eq.~\ref{eq:compat_training}). 
We observe that heterogeneous system outperforms homogeneous system on average by $3.95\%$  and $1.45\%$ for face and fashion retrieval respectively. This indicates that considering weight-compatibility alone is beneficial.

\noindent \textbf{Improvement due to architecture compatibility:}
To see the additional benefits due to architecture compatibility, in Fig.~\ref{fig:acc_break_down} we compare the heterogeneous accuracy achieved by the query models obtained via \cmpnas and trained with BCT. On IJB-C and DeepFashion2 datasets, \cmpnas outperforms the second-best model ProxyLess(Mobile) by $3.3\%$, $3.6\%$ in terms of heterogeneous accuracy. This shows architecture compatibility can improve accuracy by a large margin. Additionally, we also observe gains of $5.8\%$ and $4.8\%$ in terms of homogeneous accuracy. 

\noindent \textbf{Comparing different methods for weight-compatibility:}
In Sec.~\ref{subsec:representational_compatibility}, we discussed four ways to achieve weight-compatibility; (1) vanilla training, (2) knowledge distillation~\cite{hinton2015distilling}, (3) fine-tuning, and (4) BCT~\cite{shen2020towards}. To quantitatively compare these methods, we obtain the query network by pruning the gallery ResNet-101 model using magnitude pruning~\cite{li2016pruning} and channel pruning~\cite{he2017channel}.
To obtain the (pruned) query model, we prune $90\%$ of filters from the first two layers in each residual block of the gallery network. 
The query model obtained is trained with each of the four methods and Tab.~\ref{tab:compatibility_comparison} shows both the homogeneous and heterogeneous search accuracy.
We observe only fine-tuning and BCT can achieve weight-compatibility wherein accuracy of the heterogeneous system supersedes that of the homogeneous system. Training from scratch and knowledge distillation on the other hand, cannot ensure compatibility and obtain $0.0\%$ accuracy for heterogeneous search. 
Among fine-tuning and BCT, we prefer BCT to ensure weight compatibility for two reasons: (1) fine-tuning is restrictive: it poses a strong requirement on the query architecture \eg query model is obtained by pruning the gallery model and (2) the model trained with BCT performs better.

\begin{table}[]
\begin{adjustbox}{width=\columnwidth,center}
\Large
\begin{tabular}{ll|rrrr}
\toprule
Gallery & Query & Train & Finetune & BCT & KD \\ 
model & model & Scratch &  &  &  \\\midrule
Magnitude prune & Magnitude prune & 84.4 & 84.9 & 86.4 & 86.8 \\
ResNet-101 & Magnitude prune & 0.0 & 86.5 & 87.2 & 0.0 \\
\cdashline{1-6}[.7pt/1.5pt]\noalign{\vskip 0.15em}
Channel prune & Channel prune & 84.2 & 85.2 & 86.5 & 87.0  \\
ResNet-101 & Channel prune & 0.0 & 86.3 & 87.4 & 0.0  \\ \bottomrule
\end{tabular}
\end{adjustbox}
\caption{\small{Comparing techniques for achieving weight-level compatibility on the 1:N face retrieval task. The query model $\phi_q$ is obtain by pruning $90\%$ of filters in the first two layers of each residual block of the gallery model. We see that for both pruning methods, training the query model with BCT loss leads to the highest heterogeneous accuracy.}}
\label{tab:compatibility_comparison}
\vspace{-2mm}
\end{table}

\begin{figure}
    \centering
    \begin{subfigure}{0.48\linewidth}
    \centering
    \includegraphics[width=\linewidth]{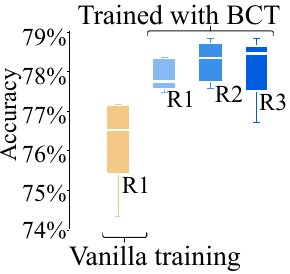}
    \caption{\small{1:N search on IJB-C.}}
    \end{subfigure}
    \begin{subfigure}{0.48\linewidth}
    \centering
    \includegraphics[width=\linewidth]{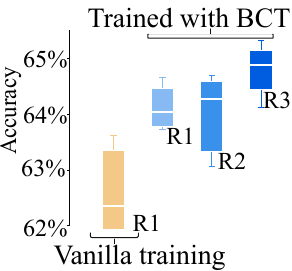}
    \caption{\small{Top-10 on DeepFashion2.}}
    \end{subfigure}
    \caption{\small{Ablating on training strategies (vanilla, BCT) and rewards ($\mathcal{R}_1-\mathcal{R}_3$) for \cmpnas 
    These plots show the heterogeneous accuracy of the best 5 models (under 100 Mflops) discovered by each method and trained from scratch with BCT. Observe that the ingredients of \cmpnas \ie BCT training + reward $\mathcal{R}_3$ perform the best.}}
    \label{fig:opt_target}
    \vspace{-5mm}
\end{figure}

\noindent \textbf{Comparing \cmpnas with baseline NAS\cite{NAS_SinglePath}:}
To measure the gains relative to baseline NAS, in Fig.~\ref{fig:opt_target}, we present a barplot of the heterogeneous accuracy achieved by the best 5 architectures obtained by vanilla NAS (yellow bar) and \cmpnas (blue bars) when trained from scratch using BCT.
The baseline considers the vanilla loss ($\lambda_2=0$ in Eq.~\ref{eq:compat_training}) to train the super-network and searches using reward $\mathcal{R}_1$ while \cmpnas uses BCT to train the super-network and can search using rewards $\mathcal{R}_1$, $\mathcal{R}_2$, $\mathcal{R}_3$. On both datasets, \cmpnas outperforms the baseline by $2-2.5\%$.

\noindent \textbf{Comparing reward choices for \cmpnas:} In Fig.~\ref{fig:opt_target}, the performance of different reward choices (of Tab.~\ref{tab:reward}) are shown by blue plots. As expected, the baseline reward  ($\mathcal{R}_1$) performs worst since its target (homogeneous accuracy) is misaligned with our target (heterogeneous accuracy). The second reward ($\mathcal{R}_2$) is much better since it directly optimizes the target while the composite reward ($\mathcal{R}_3$) works best with especially large gains observed on DeepFashion2.

\begin{table}[]
\Huge
\centering
\begin{adjustbox}{width=\columnwidth,center}
\begin{tabular}{@{}llr|rr@{}}
\toprule
Gallery  & Query & Query & Fashion retrieval & Face retrieval \\
model & model & MFlops & top-10 search & 1:N search \\ \midrule

ResNet-101 &     ResNet-101        & 7597            & 65.1               & 86.7  \\ \hline
\multirow{4}{*}{ResNet-101} 
& MobileNetV1(1x)             & 579             & 62.3                        & 73.0                \\
& MobileNetV2(1x)             & 329                  & 60.4                     & 77.0                \\
& ProxyLess(mobile)             & 332                   & 62.1                      & 81.8                \\

& CMP-NAS-a(Face)      & \textbf{327}                & 65.4              & \textbf{85.1}                \\
& CMP-NAS-a(Fashion)      & \textbf{314}                & \textbf{65.7}              & 84.2               \\ 
\cdashline{1-5}[.7pt/1.5pt]\noalign{\vskip 0.15em}
\multirow{2}{*}{ResNet-101} 
& MobileNetV3                   & 226         & 63.0                  & 80.4                   \\ 
& CMP-NAS-b(Face)          &      \textbf{216}               & 64.4     &         \textbf{84.1}            \\ 
& CMP-NAS-b(Fashion)      & \textbf{211}                & \textbf{64.9}              & 81.5                \\ 
\cdashline{1-5}[.7pt/1.5pt]\noalign{\vskip 0.15em}
\multirow{5}{*}{ResNet-101} 
& MobileNetV1(0.5x)  & 155  & 60.3  & 62.7 \\
& ShuffleNetV2(1x)   & 149  & 63.3  & 75.8 \\
& ShuffleNetV1(1x,g=1) & 148 & 62.6  & 76.0 \\
& MobileNetV2(0.5x)    & 100 & 62.0 & 73.1 \\
& CMP-NAS-c(Face)      & \textbf{94}                    & 62.4                     & \textbf{78.8} \\ 
& CMP-NAS-c(Fashion)      & \textbf{93}                & \textbf{64.8}              & 77.8                \\ \bottomrule
\end{tabular}
\end{adjustbox}
\caption{\small{Evaluating architectures searched with \cmpnas for fashion retrieval (denoted as fashion) and face retrieval (denoted as face) tasks. We search models for three different complexity tiers: 100, 230 and 330 Mflops and use the best architecture to report the results. The searched models outperform other architectures by $3\sim5\%$ on both the tasks.}}
\label{tab:compat_nas_best}
\vspace{-5mm}
\end{table}

\subsection{Generalization performance of \cmpnas}
In this section, we investigate the performance of \cmpnas under different compute constraints, application scenarios and tasks. 
Inspired by state-of-the-art architectures for mobile deployment we select three computational tiers: 330 million flops (similar to MobileNetV2),  230 Mflops (similar to MobileNetV3), and 100 Mflops (similar to ShuffleNetV2).
For each computational tier, we implement a heterogeneous system using the models searched by \cmpnas. These models are denoted by \cmpnas-a (330 Mflops), \cmpnas-b (230 Mflops), and \cmpnas-c (100 Mflops). Additionally, we append ``(Face)''/``(Fashion)'' to the model name, \eg ``\cmpnas-a(Face)''/``\cmpnas-a(Fashion)'', to denote the architecture searched on the face or fashion datasets respectively.

\noindent\textbf{\cmpnas for different resource constraints:}
We compare the performance of architectures searched by \cmpnas for each computational tier in Tab.~\ref{tab:compat_nas_best}.
For each task, we look at the model searched on the same task \eg for face retrieval we look at \cmpnas-a(Face) \etc.
On both datasets the models searched by \cmpnas consistently outperform other state-of-the-art baselines. 
For 330 Mflops tier, \cmpnas-a outperforms the second best (ProxyLess(Mobile)~\cite{proxylessnas}) by $3.6\%$ and by $3.1\%$ on the fashion and face retrieval tasks respectively. 
Similarly, \cmpnas-b outperforms the second best network MobileNetV3~\cite{MobileNetV3} by $1.9\%$ and $3.7\%$ on the corresponding tasks.
Finally, for the $100$M tier, \cmpnas-c achieves $1.5\%$ and $3.0\%$ improvement over the second best network ShuffleNetV2($1\times$) while consuming $33\%$ fewer flops. These results establish the generalization ability of the \cmpnas for different computation constraints.

\noindent\textbf{\cmpnas across different applications:}
For this experiment, we evaluate the architectures searched on the face dataset for the fashion retrieval task and vice versa. The results are shown in the Tab.~\ref{tab:compat_nas_best}.
We observe that the architectures optimized for the face tasks \cmpnas-a/b/c(Face) also outperform the baselines for the fashion retrieval task. Moreover, these models only lose $1-2\%$ accuracy compared to the best model searched on the fashion dataset. 
We make similar observations for \cmpnas-a/b/c/(Fashion) evaluated on the face retrieval task. This shows that the architectures searched by \cmpnas can generalize across application scenarios.

\noindent \textbf{\cmpnas for face verification:}
In table Tab.~\ref{tab:compat_nas_face_verification}, we use the \cmpnas-a/b/c(Face) models for another ``open universe'' problem: 1:1 face verification. The results show that the models searched by \cmpnas outperform state-of-the art architectures by $3-5\%$ in the homogeneous and heterogeneous settings. Importantly, the compatibility rule is also achieved. This indicates the searched models can generalize across different tasks.

\begin{table}[]
\centering
\footnotesize
\begin{adjustbox}{width=\columnwidth,center}
\begin{tabular}{@{}llr|rr@{}}
\toprule
Gallery & Query  & Query   & Homogeneous & Heterogeneous     \\
model& model &  MFlops  & accuracy & accuracy \\ \midrule
                               
ResNet-101&     ResNet-101        & 7597                         & 85.4 & -  \\ \hline

\multirow{2}{*}{ResNet-101} &  ProxyLess(mobile)             & 332   &  75.5 & 80.3        \\
 &CMP-NAS-a(Face)      & \textbf{327}   &  \textbf{81.6}    &   \textbf{84.5}          \\ 
\cdashline{1-5}[.7pt/1.5pt]\noalign{\vskip 0.15em}

\multirow{2}{*}{ResNet-101} & MobileNetV3                   & 226    & 74.3 & 79.9            \\ 
 &CMP-NAS-b(Face)          &  \textbf{216} &  \textbf{79.0} & \textbf{82.8}          \\
\cdashline{1-5}[.7pt/1.5pt]\noalign{\vskip 0.15em}

\multirow{2}{*}{ResNet-101} & ShuffleNetV2(1x)   & 149   & 66.8 & 74.8\\
 &CMP-NAS-c(Face)      & \textbf{94} & \textbf{71.5} & \textbf{78.3} \\  

\bottomrule
\end{tabular}
\end{adjustbox}
\caption{\small{Evaluating the models \cmpnas-a,b,c(Face) on the 1:1 face verification task using IJB-C. Accuracy metric is \footnotesize{TAR\text{@}FAR=$10^{-4}$}}. The searched models outperform the baselines indicating they can generalize across tasks. }
\label{tab:compat_nas_face_verification}
\vspace{-5mm}
\end{table}
\section{Discussion}
We have presented a heterogeneous visual search system that achieves high accuracy with low computational cost.
Key to building this system is ensuring the query and gallery models are compatible. 
We achieve this through joint weight and architecture compatibility optimization with \cmpnas. 
There are, however, some limitations of our method: 
(1) Our method is limited to classification-based embedding training and does not directly work with metric learning based approaches;
(2) We consider the simplified use-case for architecture optimization wherein the gallery model is fixed. 
A more general treatment of model compatibility may optimize both the gallery and query models.
These limitations show that there is scope for improving our \hevs system which can be tackled by future work.

{\small
\bibliographystyle{ieee_fullname}
\bibliography{reference.bib}
}

\clearpage
\section*{Appendices}
\label{sec:appendix}
\appendix

\section{Implementation Details}
\label{Sec:ImplementationDetails}
\subsection{Training, Validation and Testing Dataset}

For searching the best query architecture, we carve out a small validation split from the original training set. For the face tasks, we set aside $5\%$ from the training set of IMDB \cite{IMDB} while for the fashion tasks, we set aside $10\%$ from the training set of DeepFashion2 \cite{Deepfashion2}. The remaining portions of the training sets are actually used to train all our embedding models (query supernet, gallery model, final query models). After a super-network is trained, we evaluate the performance of each candidate architecture (we refer to it as a sub-network) on the held out validation split. The final results presented in this paper are reported on the original validation portions of IMDB and DeepFashion2.

\subsection{Designing and Training the Super-network}
For each computational tier (330, 230, 100 Mflops), we train a different super-network. 
For the 300 Mflops tier, our super-network is the same as that in \cite{NAS_SinglePath}. 
For the 230 and 100 Mflops tiers, we reduce the channel widths by $0.75\times$ and $0.5\times$ in each layer. 
The super-network is trained through a sampling process: In each batch, a new architecture (we call this a sub-network) is sampled and only the weights corresponding to it are updated. 
For sampling a sub-network, we use the parameter free \textit{uniform sampling} method. 
This means that, for each layer, the chosen block (includes four choices from 0-3) and channels width (includes ten choices from 0-9) are sampled uniformly. 
We notice that the super-network fails to converge if the sampling process is started from the first epoch. 
To solve this, we use a warm-up phase of 10 epochs wherein the the super-network is trained without sampling. 
During the warm-up phase, the output of all four blocks in each layer are combined through averaging and the largest channel width is used.

\subsection{Details of the evolutionary search}
 We reuse the same hyper-parameters from \cite{NAS_SinglePath} for the evolutionary search step. 
 Specifically, we search for 20 generations, each with a population size of 50, crossover size of 40, mutate chance of 0.1 and random select chance of 0.1. 
 To guide the evolutionary search for finding the most compatible architectures, we use reward $\mathcal{R}_3$ from Tab.1 in the main paper. 
 For the face tasks, we compute this reward on the IMDB ``validation'' split using the 1:1 verification metric of TAR\text{@}FAR=$10^{-3}$.
 For the fashion tasks, we compute this reward on the DeepFashion2 validation split using the top-50 metric.
 Note that our rewards metrics (TAR\text{@}FAR=$10^{-3}$ for face, top-50 for fashion) are different from the target metrics (TAR\text{@}FAR=$10^{-4}$/TNIR\text{@}FPIR=$10^{-1}$ for face, top-10 for fashion). This is mainly because the validation split is smaller (than the test split), and thus target metric (\eg top-10 accuracy) is noisy compared to the validation metric (\eg top-50 accuracy).

\section{Additional Results under Different Evaluation Metrics}
\label{Sec:MoreResults}
Due to space limits, in the main paper, we present one evaluation metric per task.
In this section, we present the full metric results according to IJB-series and DeepFashion2 benchmark standard for reference.
More specifically, in Sec.~\ref{subsec:FaceRetrieval} we show top-k search accuracy on face retrieval task. 
In Sec~\ref{subsec:FaceVer}, we evaluate our \cmpnas on face verification task at additional operating points; 
In Sec~\ref{subsec:fashionretrieval}, we show the results of the proposed method using top-1, top-10 and top-20 retrieval accuracy on fashion retrieval task.
All these additional results further demonstrate that (1) With \cmpnas, the compatibility rule holds; (2) The architectures searched with \cmpnas outperform other baselines for both homogeneous and heterogeneous search accuracy.

\subsection{Additional Results on Face Retrieval}
\label{subsec:FaceRetrieval}

Tab.~\ref{appendix:face_ret} extends Tab.5 in the main paper by including other popular metrics (top-1, top-5 and top-10) for the face retrieval task. Additionally, we include the homogeneous accuracy achieved by the models.

\begin{table}[]
\begin{adjustbox}{width=\columnwidth,center}
\Large
\begin{tabular}{lcccccccc}
\toprule
Query Model       & MFlops &\multicolumn{3}{c}{Homogeneous Acc.} & &\multicolumn{3}{c}{Heterogeneous Acc} \\
&       &\multicolumn{3}{c}{Top-k with k=}    &       & \multicolumn{3}{c}{Top-k with k=}             \\ \cline{3-5} \cline{7-9}
                  &       &1           & 5    & 10   &  & 1            & 5       &   10   \\ \midrule
ResNet-101        &  7597   &     91.1        &     95.0          &  96.1 &                   &    -    &    -    &   -\\ \hline
MobileNetV1       & 579 &80.0             & 88.9      &   91.5 &   & 83.5             & 91.4       &   93.7   \\
MobileNetV2       & 329 &85.8             & 92.2       &  94.2 &  & 88.1             & 93.8         &   95.2 \\
ProxylessNAS      & 332 &86.3             & 92.5       &  94.4&   & 88.5             & 93.9         &  95.4  \\
CMP-NAS-a(Face)   & \textbf{327} & \textbf{89.7}      &    \textbf{94.2}   &  \textbf{95.5 }&      & \textbf{90.7}             & \textbf{94.7 }      &  \textbf{96.1}    \\
\cdashline{1-9}[.7pt/1.5pt]\noalign{\vskip 0.15em}
MobileNetV3       & 226 &85.6             & 92.1      & 94.0  &   & 88.0             & 93.5   & 95.2          \\
CMP-NAS-b(Face)   & \textbf{216} & \textbf{88.2}       &      \textbf{93.5}   &  \textbf{95.2}&       & \textbf{89.8}             & \textbf{94.5}      &     \textbf{95.9}  \\
\cdashline{1-9}[.7pt/1.5pt]\noalign{\vskip 0.15em}
MobileNetV1(0.5x) & 155 & 74.1             & 77.5       &  85.3 &  &    77.5          & 88.3     &   91.3     \\
ShuffleNetV2      & 149 &81.6             & 89.8      &  92.2  &  & 85.0             & 92.0      &   94.1   \\
ShuffleNetV1(g=1) & 148 &81.3             & 89.7      &  92.1  &  & 85.1             & 92.1      &   94.0   \\
MobileNetV2(0.5x) & 100 &80.0             & 88.5      &  91.3  &  & 83.6             & 90.9      &   93.3    \\
CMP-NAS-c(Face)   & \textbf{94}  &\textbf{84.3}       &  \textbf{91.4}      & \textbf{93.4}  &  & \textbf{86.9}             & \textbf{93.1}      &   \textbf{94.9}   \\ \bottomrule
\end{tabular}
\end{adjustbox}
\caption{\small{Extending Tab. 5 of the main paper. Evaluating \cmpnas on the IJB-C 1:N face retrieval benchmark using two additional metrics: top-1, top-5 top-10 accuracy. Observe that the models discovered with \cmpnas comprehensively outperform the baselines on both, homogeneous and heterogeneous accuracy.}}
\label{appendix:face_ret}
\end{table}

\subsection{Additional Results on Face Verification}
\label{subsec:FaceVer}
Besides face retrieval, face verification is another popular task in the ``open-universal'' problem of face recognition.
in Tab.~\ref{appendix:face_ver}, we extend Tab. 6 of the main paper by showing the results on additional operating points (FAR=$10^{-2}, 10^{-3},10^{-4}$).

\begin{table}[]
\Huge
\centering
\begin{adjustbox}{width=\columnwidth,center}
\begin{tabular}{@{}llccccccc@{}}
\toprule
Query Model                 & MFlops & \multicolumn{3}{c}{Homogeneous Acc.} & &\multicolumn{3}{c}{Heterogeneous Acc.} \\
                            &        & \multicolumn{3}{c}{TAR\text{@}FAR=}   & &\multicolumn{3}{c}{TAR\text{@}FAR=}    \\ 
                             \cline{3-5} \cline{7-9}
                               &    & $10^{-2}$        & $10^{-3}$      & $10^{-4}$    &   & $10^{-2}$   & $10^{-3}$   & $10^{-4}$                \\ \midrule
                               
Resnet-101(gallery)             & 7597       & 96.9        & 92.8      & 85.4             &      & -    &   -  &     -                         \\ \hline

MobileNetV1(1x)             & 579           & 93.2      & 82.6     & 66.7          &            &  95.0 &  86.6  & 73.0                               \\
MobileNetV2(1x)             & 329        & 95.6      & 88.1     & 75.4             &         & 96.5  &  91.0  & 80.8                              \\
ProxyLess(mobile)             & 332       & 95.7    & 88.2    & 75.5              &          & 96.5  &  90.7  & 80.3                               \\

CMP-NAS-a(Face)      & \textbf{327}     & \textbf{96.7}       & \textbf{91.5}      & \textbf{81.6 }            &           &   \textbf{97.1}    &   \textbf{92.7} & \textbf{84.5}                                 \\ \hline

MobileNetV3                   & 226     & 95.5       &      88.0      & 74.3                   &      &   96.5  &  90.9   & 79.9                                       \\ 

CMP-NAS-b(Face)          &      \textbf{216}       & \textbf{96.3}     & \textbf{90.2}     &  \textbf{79.0}           &      & \textbf{96.9}               &   \textbf{92.2}   & \textbf{82.8}                             \\  \hline
MobileNetV1(0.5x)           & 155      & 90.8       & 76.9       & 58.0             &         & 93.4          &  82.1     & 64.3                                     \\
ShuffleNetV2(1x)             & 149       & 93.7        & 83.8      & 66.8       &       & 95.4               &    88.7    & 74.8                                 \\
MobileNetV2(0.5x)            & 100       & 93.3       & 82.0       & 64.8        &         & 94.9              &    86.8   & 72.8                                    \\

CMP-NAS-c(Face) & \textbf{94 }      & \textbf{95.1}     & \textbf{86.6}        & \textbf{71.5 }     &         & \textbf{96.1}            &    \textbf{90.2}   & \textbf{78.3}                          \\  \bottomrule

\end{tabular}
\end{adjustbox}
\caption{\small{Extending Tab. 6 of the main paper. Evaluating the models \cmpnas-a,b,c(Face) on the 1:1 face verification task using IJB-C using additional operating points.  The searched models outperform the baselines indicating they can generalize across tasks.}}
\label{appendix:face_ver}
\end{table}

\subsection{Additional Results on Fashion Retrieval}
\label{subsec:fashionretrieval}
Tab.~\ref{appendix:fashion_ret} extends Tab. 5 in our paper by showing the homogeneous and heterogeneous accuracy through the top-1, top-10 and top-20 metrics.

\begin{table}[]
\begin{adjustbox}{width=\columnwidth,center}
\Huge
\begin{tabular}{lllccccccc}
\toprule
Query Model       & MFlops & \multicolumn{3}{c}{Homogeneous Acc.} & &\multicolumn{3}{c}{Heterogeneous Acc} \\
                  &       & \multicolumn{3}{c}{Top-k with k as}  & &\multicolumn{3}{c}{Top-k with k as}   \\
                  \cline{3-5} \cline{7-9} 
                  &       & 1   & 10  & 20    & & 1   & 10  & 20    \\ \midrule
ResNet-101        &        &    39.4     &       65.1          &   72.0 &     &  -                &     -     &   -     \\ \hline
MobileNetV1       &   579  &    34.7     &       60.5          &   67.8 &     &  36.3                &     62.3     &   69.1    \\
MobileNetV2       &   329  &    32.4     &       58.0          &   65.9 &     &  33.9                &     60.4     &   67.9     \\
ProxylessNAS      &   332  &    35.1     &       60.8          &   68.5 &     &  36.6                &     62.1     &   69.4     \\
CMP-NAS-a(Fashion)&   \textbf{314}  &    \textbf{39.0}     &       \textbf{65.4}          &   \textbf{72.4} &     &  \textbf{39.3}                &     \textbf{65.6}     &   \textbf{72.5}    \\
\cdashline{1-9}[.7pt/1.5pt]\noalign{\vskip 0.15em}
MobileNetV3       &   226  &    37.1     &       62.7          &   69.9 &     &  37.5                &     63.0     &   70.2     \\
CMP-NAS-b(Fashion)&   \textbf{211}  &    \textbf{38.2}     &       \textbf{64.0}          &   \textbf{71.2} &     &  \textbf{38.4}                &     \textbf{64.9}     &   \textbf{72.2}    \\
\cdashline{1-9}[.7pt/1.5pt]\noalign{\vskip 0.15em}
MobileNetV1(0.5x) &   155  &    32.8     &       57.7          &   65.6 &     &  34.0                &     60.2     &   67.5     \\
ShuffleNetV2      &   149  &    35.4     &       60.7          &   68.1 &     &  35.7               &     62.1     &   69.7     \\
ShuffleNetV1(g=1) &   148  &    34.4     &       60.5          &   68.1 &     &  35.3                &     62.6     &   69.8     \\
CMP-NAS-c(Fashion)   &   \textbf{93}   &    \textbf{37.6}     &       \textbf{63.5}          &   \textbf{71.0} &     & \textbf{ 38.4}                &     \textbf{64.8}     &   \textbf{72.1}   \\ \bottomrule   
\end{tabular}
\end{adjustbox}
\caption{\small{Extending Tab. 5 of the main paper. Evaluating \cmpnas on the Deepfashion2 fashion retrieval benchmark using additional metrics: top-1 and top-20 accuracy. We observe that the models discovered with \cmpnas comprehensively outperform the baselines on both, homogeneous and heterogeneous accuracies.}}
\label{appendix:fashion_ret}
\end{table}

\section{Additional results for weight-level compatibility}
\label{sec:TrainingMethod}
Due to space limit, Tab 4 in the main paper compares different training methods for weight-level compatibility using query model achieved by pruning $90\%$ of filters.
Tab.~\ref{appendix:pruning} extends Tab 4 in the main paper by showing heterogeneous accuracy of query models achieved by pruning the gallery model to different levels.

\begin{table}[]
\begin{adjustbox}{width=\columnwidth,center}
\Large
\begin{tabular}{llc|rrrr}
\toprule
Gallery & Query Prune& Prune &Train & Fine-tune & BCT & KD \\ 
model & method & Amt. &Scratch &  &  &  \\\midrule
ResNet-101 & -  &  0\%& 87.9& -& - & -\\ \hline
ResNet-101 & Magnitude \cite{li2016pruning}  &  30\%& 0.0 & 87.9& \textbf{88.5} & 0.0\\
ResNet-101 & Magnitude \cite{li2016pruning} &  50\%& 0.0& 87.3& \textbf{88.2}& 0.0\\
ResNet-101 & Magnitude \cite{li2016pruning}  &  70\%& 0.0& 87.2& \textbf{87.9}& 0.0\\
ResNet-101 & Magnitude \cite{li2016pruning}  &  90\%&0.0 & 86.5 & \textbf{87.2} & 0.0 \\
\cdashline{1-7}[.7pt/1.5pt]\noalign{\vskip 0.15em}
ResNet-101 & Channel \cite{he2017channel}  &  30\%& 0.0 & 87.6 & \textbf{88.4} & 0.0\\
ResNet-101 & Channel \cite{he2017channel}  &  50\%& 0.0& 87.5& \textbf{87.8} & 0.0\\
ResNet-101 & Channel \cite{he2017channel}  &  70\%& 0.0 & 87.3 & \textbf{87.9}& 0.0\\
ResNet-101 & Channel \cite{he2017channel}  & 90\% & 0.0 & 86.3 & \textbf{87.4} & 0.0  \\  \bottomrule
\end{tabular}
\end{adjustbox}
\caption{\small{Extending Tab. 4 of the main paper. Comparing training methods for heterogeneous accuracy achieved on the 1:N face retrieval task. The query model $\phi_q$ is obtained via pruning filters from the first two layers of each residual block of the gallery model. We compare two different pruning methods \cite{he2017channel,li2016pruning} at several pruning amounts. Observe that for all pruning methods and amounts, training the query model with BCT loss leads to (1) non-zero heterogeneous accuracy and (2) the highest heterogeneous accuracy.}}
\label{appendix:pruning}
\end{table}

\begin{figure*}[t!]
    \centering
    \begin{subfigure}{0.24\linewidth}
    \centering
    \includegraphics[width=0.8\linewidth]{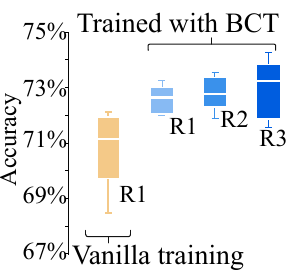}
    \caption{\small{Homogeneous (Face)}}
    \end{subfigure}
    \begin{subfigure}{0.24\linewidth}
    \centering
    \includegraphics[width=0.8\linewidth]{Images/ablation_rewards_face.pdf}
    \caption{\small{Heterogeneous (Face)}}
    \end{subfigure}
     \begin{subfigure}{0.24\linewidth}
    \centering
    \includegraphics[width=0.8\linewidth]{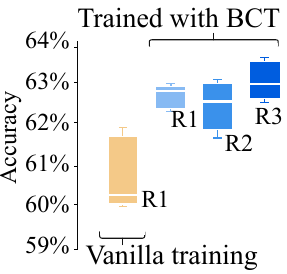}
    \caption{\small{Homogeneous (Fashion)}}
    \end{subfigure}
    \begin{subfigure}{0.24\linewidth}
    \centering
    \includegraphics[width=0.8\linewidth]{Images/ablation_rewards_fashion.pdf}
    \caption{\small{Heterogeneous (Fashion)}}
    \end{subfigure}
    \caption{\small{Extending Fig.~6 of the main paper. The figures are generated by averaging the best five architectures discovered by \cmpnas (under 100 Mflops) when using different training strategies (Vanilla, BCT) and rewards ($\mathcal{R}_1-\mathcal{R}_3$). In (a),(b) we plot the homogeneous and heterogeneous accuracy for the 1:N face retrieval task using the metric TNIR\text{@}FPIR=$10^{-1}$. In (c),(d) we plot the homogeneous and heterogeneous accuracy for the fashion retrieval task using the metric top-10. Observe that in all cases, BCT training works best among the training strategies while $\mathcal{R}_3$ outperforms all other rewards.}}
    \label{appendix:rewards}
\end{figure*}

\section{Comparing different rewards}
\label{sec:rewards}
Fig.~\ref{appendix:rewards} is an extension of Fig.~6 in the paper.
We present the homogeneous and heterogeneous accuracy achieved by the best five query models searched using different rewards and training schemes on the face and fashion retrieval tasks.
These complementary results further reinforce our conclusions: $\mathcal{R}_3$ generally works better than $\mathcal{R}_1$ and $\mathcal{R}_2$; Training the super-network with BCT outperforms vanilla training by a large margin.


\end{document}